\documentclass[letterpaper, 10 pt, conference]{ieeeconf}  % Comment this line out if you need a4paper

\IEEEoverridecommandlockouts                              % This command is only needed if 
\usepackage{graphics} % for pdf, bitmapped graphics files

\usepackage{cite}
\usepackage{amsmath,amssymb,amsfonts}
\usepackage{algorithmic}
\usepackage{graphicx}
\usepackage{textcomp}
\usepackage{xcolor}
\usepackage{comment}
\usepackage{hyperref}
\usepackage{bm}
\usepackage{wrapfig}
\usepackage{xcolor}
\usepackage{multirow}
\usepackage{subfig}
\newcommand\note[1]{}    
\usepackage{float}
\usepackage{booktabs}

\title{\LARGE \bf
Generating and Characterizing Scenarios for Safety Testing of Autonomous Vehicles}

\author{ %
 Zahra Ghodsi$^1$, 
 Siva Kumar Sastry Hari$^2$, 
 Iuri Frosio$^2$, 
 Timothy Tsai$^2$,
 Alejandro Troccoli$^2$, \\
 Stephen W. Keckler$^2$, 
 Siddharth Garg$^1$, and
 Anima Anandkumar$^2$ \\
$^1$New York University
$^2$NVIDIA Corporation
}
\begin{document}

\maketitle
                                                          \thispagestyle{empty}
\pagestyle{empty}

%%%%%%%%%%%%%%%%%%%%%%%%%%%%%%%%%%%%%%%%%%%%%%%%%%%%%%%%%%%%%%%%%%%%%%%%%%%%%%%%
\begin{abstract}
Extracting interesting scenarios from real-world data as well as generating failure cases is important for the development and
testing of autonomous systems. We propose efficient mechanisms to both characterize and generate testing scenarios using a state-of-the-art driving simulator.
For any scenario, our method generates a set of possible driving paths and identifies all the possible safe driving trajectories that can be taken starting at different times, 
to compute metrics that quantify the complexity of the scenario. We use our method to characterize real driving data from the Next Generation Simulation (NGSIM) project, as well as adversarial scenarios generated in simulation.
We rank the scenarios by defining metrics based on the complexity of avoiding accidents and provide insights into how the AV could have minimized the probability of incurring an accident. We demonstrate a strong correlation between the proposed  metrics and human intuition.

\end{abstract}

\section{Introduction}
Research and development of Autonomous Vehicles (AV) has 
surged in recent years thanks to advances in 
perception and planning% due to the use of machine learning
, with many companies starting on-the-road testing\cite{avtest}. 
Despite these breakthroughs, 
a successful uptake of this technology remains challenging due to safety concerns and its complexity.
While AVs are being tested in the real world, comprehensive testing requires 
prohibitive driving times because the occurrence of unsafe conditions is 
rare~\cite{kalra2016driving}. Such testing is also expensive and hazardous. 
Fast methods to uncover unsafe driving conditions that are challenging 
for the AV systems at a reasonable cost are therefore valuable for making AVs safe.

Physics-based simulators enable comprehensive testing of the AV under
conditions similar to the real world, and photo-realistic simulators allow
testing of the perception modules in the AV stack. Significant advancements
have been made in developing such simulators~\cite{DriveSim, carla, lgsvl}
providing a scalable and safe environment for testing AVs.
Custom scenarios and corner cases can be tested in simulation under 
various conditions that are often hard to encounter in reality but are 
valuable for safety testing.
Prior research proposed methods to generate scenarios for safety testing 
of AVs~\cite{corso2019adaptive, abeysirigoonawardena2019generating}. 
Potentially unsafe scenarios are generated  using reinforcement
learning or optimization methods that aim to reduce the distance between the
AV and an attacker (e.g., car or pedestrian).\note{add more} 
In these methods, a scenario is considered useful if it involves a 
collision or a near collision. However, not all scenarios that end in an 
accident are useful because the AV may not be able to avoid
the accident given minimum required times for object perception, 
processing, and actuator command issuance by the AV software and for 
the car to execute the evasive maneuver and physically move. 

Instead, we consider a scenario as \emph{interesting} if the AV can avoid the collision in at least one way. 
Testing the AV using this scenario allows us to identify shortcomings and bugs in the AV system that when addressed, can avoid the collision. 
Additionally, we need the ability to characterize the generated scenarios based on 
the varying levels of difficulty (possibility of avoiding the collision) 
such that the scenarios can be prioritized for AV debugging or for learning methods.

\begin{figure}
    \centering
    %\subfloat[]{\includegraphics[height=0.2\textwidth]{./figs/teaser_scenario_narrow.png}\label{fig:scenario}}
    \subfloat[]{\includegraphics[height=0.17\textwidth]{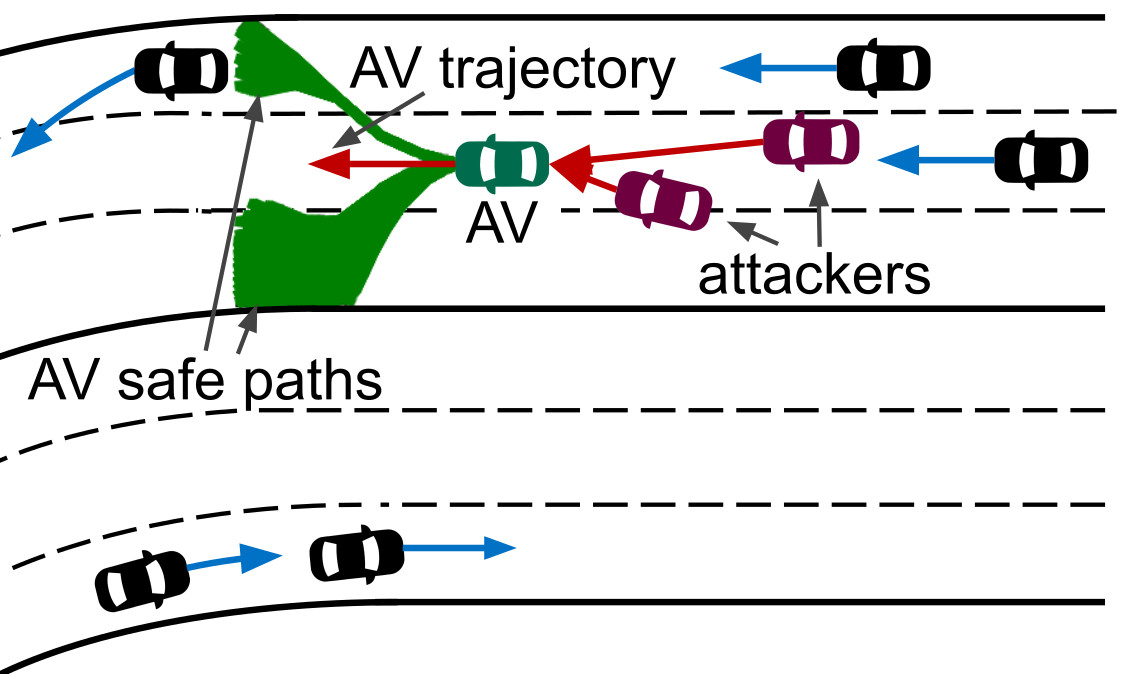}\label{fig:scenario}}
    %\hspace{1cm}
    %\subfloat[]{\includegraphics[width=0.25\textwidth]{./figs/teaser_radar.png}\label{fig:radar}}
    \subfloat[]{\includegraphics[width=0.2\textwidth]{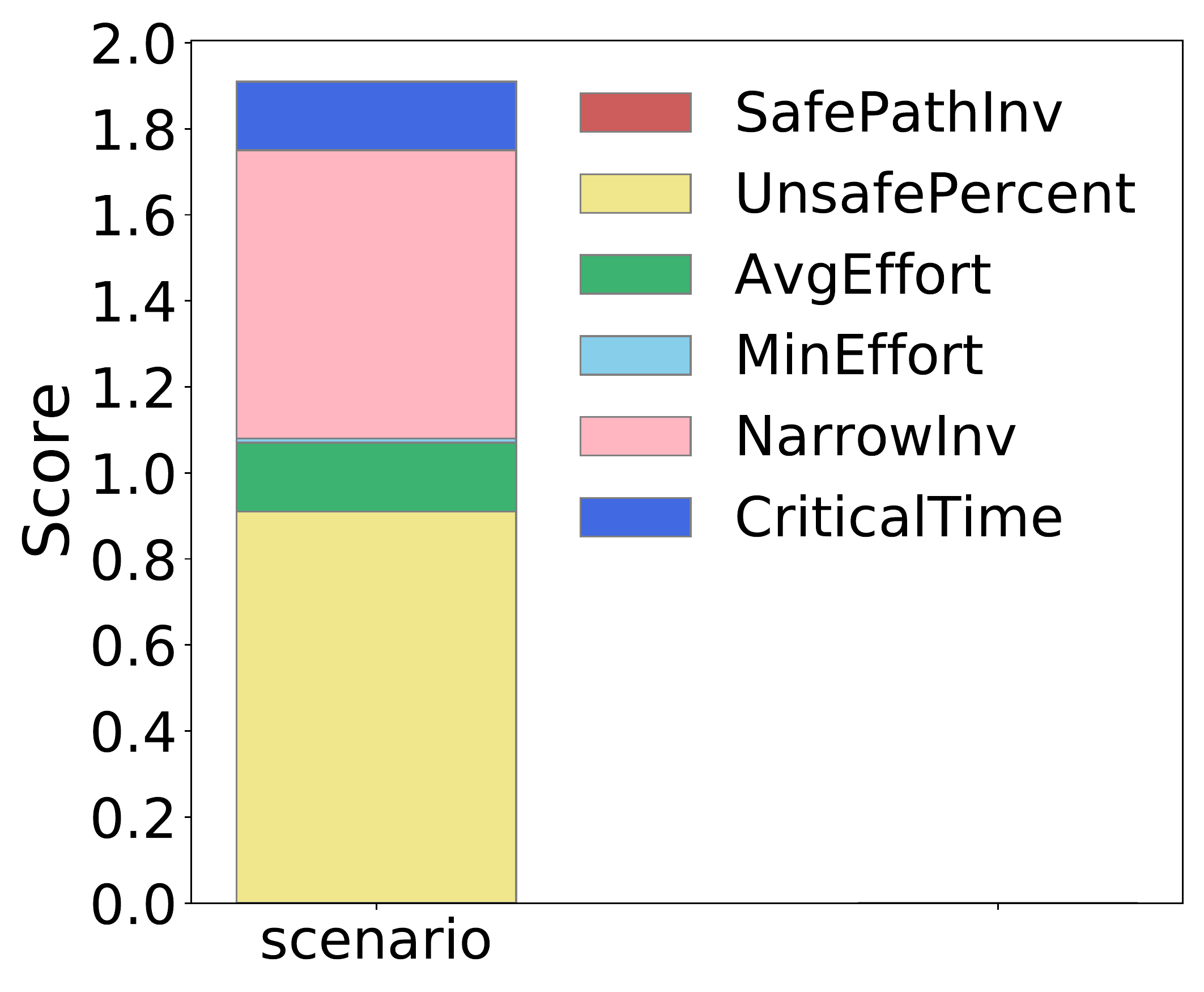}\label{fig:radar}}
    \caption{(a) depicts a generated unsafe scenario, with trajectories for all vehicles from 1.5s before to 1s after collision. The base trajectory of the AV (resulted in collision) as well as safe driving paths found are shown. The breakdown of characterization metrics for this scenario is shown in (b).}
\label{fig:teaser}
\end{figure}

The contributions of this work are twofold. 
First, we develop a method to characterize driving scenarios used for safety 
testing of AVs. To the best of our knowledge, characterizing unsafe 
scenarios based on the avoidability of the accident has never been explored. 
We define metrics to characterize scenarios based on 
the following factors for the AV: the number of safe driving paths, 
total paths in the scenario, narrowness of safe paths, and the effort 
required to follow each safe path.
We enumerate and store the set of paths using a computationally 
efficient tensor representation and calculate the above metrics in 
the tensor space. We show that our characterization method can extract 
interesting scenarios from real driving data~\cite{ngsim}.

Second, we develop a method for generating unsafe scenarios which can take an 
initial executed driving scenario as input (generated randomly, from traffic 
models, or even from real data) and introduce perturbations such that the 
likelihood of an unsafe condition increases. Our method models dynamic 
actors who act as attackers (or distracted drivers) for a fixed time with 
constraints on acceleration and steering. The attacker aims to create an 
unsafe condition by reducing the distance between itself and the AV. 
Our proposed scenario generation method can create a variety of scenarios 
with varying levels of difficulty using multiple modes for generating a 
scenario with constraints on acceleration and steering controls of the attacker 
actor. This method is not constrained to a specific AV policy and can dynamically 
generate unsafe scenarios for any AV under test.
We generate approximately $240$ scenarios per hour on a single system with 
up to $80$ accidents using our method. Results suggest that more than 
$90\%$ of the generated accidents are avoidable with appropriate actions 
taken (by an ideal AV system) approximately 2 seconds before the collision.

Fig.~\ref{fig:teaser} shows the typical output of our scenario generation and
characterization method. In this example, two attackers try to 
hit the AV from behind. The computed safe paths are shown in green in 
Fig.~\ref{fig:scenario}, which suggest that changing the lane after detecting 
the accelerating rear vehicles can avoid the accident. 
The breakdown of our metrics is shown in Fig.~\ref{fig:radar}, where UnsafePercent represents the percentage of trajectories available to the AV that are unsafe, and NarrowInv represents the narrowness of a safe trajectory and the precision required to navigate through it.
The large UnsafePercent indicates that most of maneuvers from the AV inevitably lead to a collision, and only a few trajectories lead to changing the lane and a safe outcome. A large NarrowInv also indicates that the AV has to take a precise action, i.e., change lane at the very beginning of the scenario, to avoid collision.
If the AV is not designed to take this action, the information from this scenario can be used to improve the driving policy.
Similarly, our characterization method for different scenarios can help developers gain a better understanding of the AV performance and prioritize these scenarios in AV policy development and improvement.

\section{Generating and Characterizing Scenarios}\label{method}
We define a \emph{driving scenario} such that it includes the description 
of the environment, number of actors, initial states of the actors, and 
the driving policies 
of all actors except for the AV. Such a definition allows using a scenario to test or benchmark different AV policies. A driving scenario includes the following.
    (1) A set of descriptors to characterize the environment, e.g., the number of lanes, their widths, and the road curvature.
    (2) A set of $N$ vehicles $\{w_i\}_{i=1..N}$ and their states $\{\mathbf{s}_i(t_0)\}_{i=1..N}$ at time $t_0$; the state includes the vehicle position $\mathbf{x_i}$, orientation $\theta_i$, and speed $v_i$; the size of each vehicle is also provided.
    (3) A set of driving policies $\{\pi_i\}_{i=1..N}$ associated with each non-AV vehicle.
    (4) One AV indicated with $w_{AV}$ and its state at time $t_0$, $\mathbf{s_{AV}}(t_0)$.

We use the term \emph{sequence} to indicate a temporal succession of states of all the vehicles.
Therefore, in a sequence the trajectories of the vehicles are fixed.
Sequences can be extracted from real-world data or from driving simulations. 
To create a scenario, we can extract the state of a set of vehicles (including the AV) at time $t$ in a sequence.
We initialize the state of all the vehicles in the scenario with these sampled states, and assign policies to all of them but the AV.
A simple policy is to follow the trajectories that have been measured in the sequence.  More complicated policy implementations such as those described in Section~\ref{sec:gen} can also be considered.
We point out that not all sampling times result in interesting scenarios, e.g., the corresponding scenario can be easy to solve
when the AV is far from the other cars.
As stated in the Introduction, we instead look for interesting scenarios, where a collision or near collision happens, and the AV can avoid it in at least one way. This is because if the collision is unavoidable, there is no solution independent of the AV policy.
We sample interesting scenarios by starting from collision time $t_c$ and moving back in time to find $t_0$ such that the AV can follow at least one safe path to avoid the collision (more detail in Section~\ref{sec:characterize}).
It is worth noting that our characterization method can characterize \emph{any} scenario, and can also be used to identify interesting or diverse scenarios.

\subsection{Characterization method}\label{sec:characterize}

\begin{figure}
\centering
\includegraphics[width=0.42\textwidth]{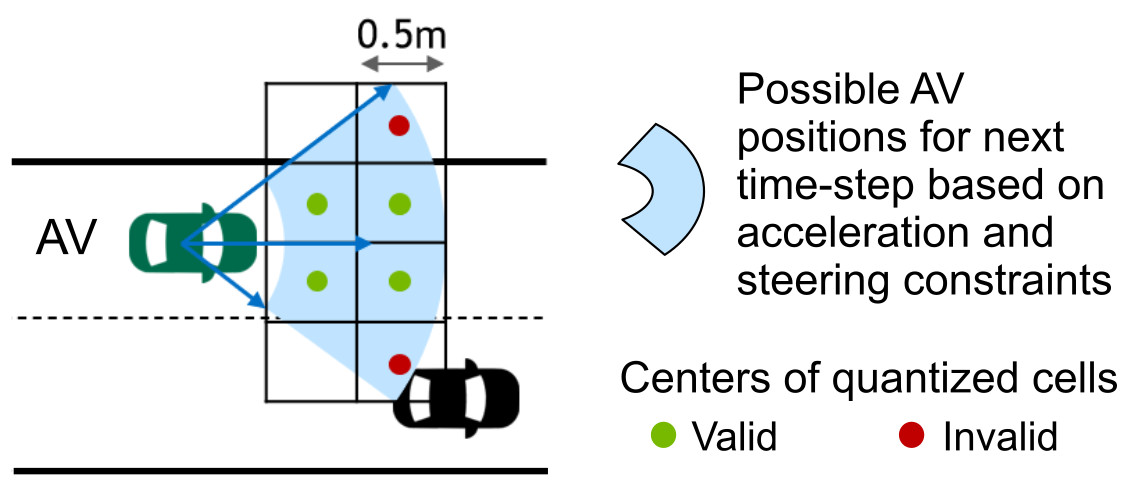}
\caption{The quantized region on the map where the AV can travel in the next time-step based on maximum allowed steering and acceleration}
\label{fig:quant}
\end{figure}

\begin{figure*}[t]
    \centering
    \subfloat[]{\includegraphics[width=0.7\textwidth]{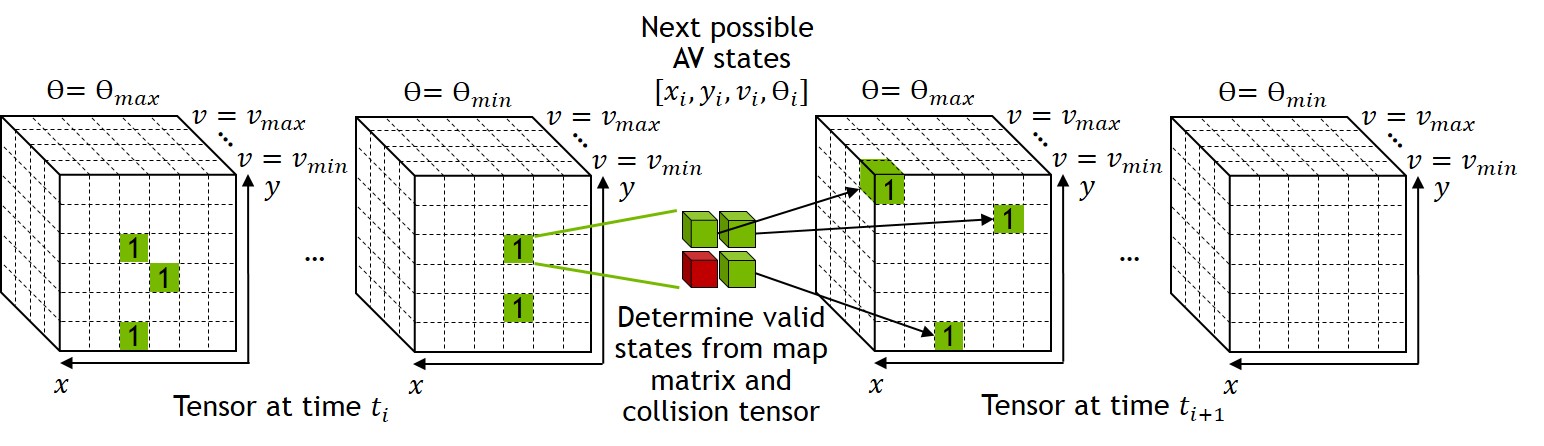}}
    \hfill
    \subfloat[]{\includegraphics[width=0.3\textwidth]{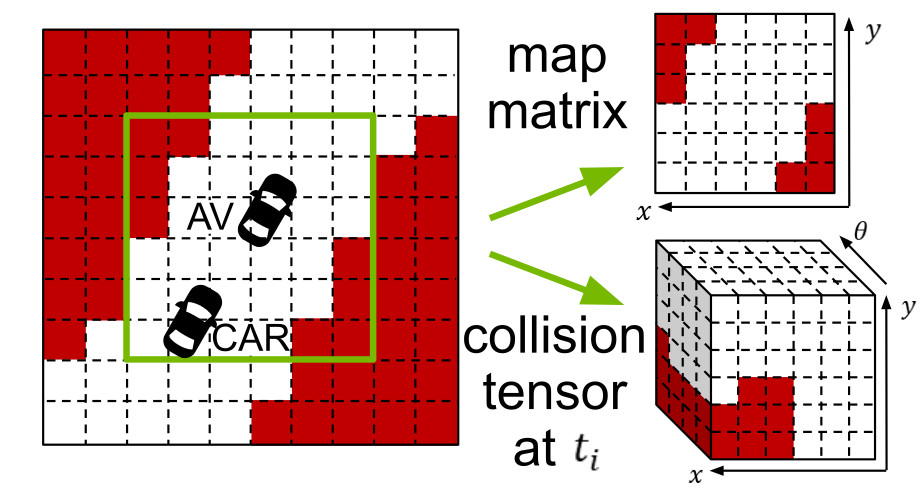}}
    \caption{Scenario scoring based on search for safe driving policy. (a) represents the propagation of tensor representing AV state from time $t_0$ to time $t_1$, by computing next possible AV states for each valid state at $t_0$, and determining next valid states based on map matrix and collision tensor. (b) depicts pre-calculation of map matrix and collision tensor from a scenario.}
    \label{fig:tensor}
\end{figure*}

Once a scenario is defined, we want to characterize how difficult it is to be solved by the AV, i.e. how many safe paths exist and how hard is it to follow them.
A naive approach to find a safe path for the AV in a scenario would be to run simulations with different driving policies, which would incur unacceptable runtime overheads.
Instead, starting from a given $t_0$ 
(e.g., three time-steps before the sequence end time $t_{end}$), we calculate a set of possible locations of the AV in the next time-step.
We compute the possible locations by calculating the annulus sector (shown in light blue color in Fig.~\ref{fig:quant}) corresponding to all the possible locations for the center of the vehicle, based on the maximum possible acceleration (positive or negative) and steering of the AV at its current state.
We discretize the 2D sector with a grid with 0.5m spacing between the cells (different spacing can also be used). 

An approach to compute all safe paths would be to build a tree with all the possible trajectories that can be executed by the AV (starting at $t_0$). Each node in the tree at depth $i$ corresponds to a valid state of AV at time-step $t_i$ defined by AV coordinates $(x,y)$, orientation $\theta$, and velocity $v$. Each node stores the corresponding AV state along with the state of other actors in the scenario at $t_i$. The children of each node are computed by recording the state of the AV for cells whose centers fall into the annulus in Fig.~\ref{fig:quant}. The cells whose centers are either off-road or result in a collision with another vehicle correspond to invalid AV states and are discarded. Valid state coordinates and invalid state coordinates for AV are shown as green and red dots in Fig.~\ref{fig:quant} respectively. For each new valid AV state, we create a node in the tree and iterate the procedure until we reach the time-step $t_{end}$.
In this structure, a safe trajectory is equivalent to a path from the root to a leaf in the tree (all tree nodes correspond to valid AV states).
If no safe path exists, we would move back one time-step from $t_0$ and repeat 
the procedure until a safe path is found.
This approach could also be used to find a \emph{critical scenario}, 
i.e., a scenario beginning from time-step $t_0$ that 
results in a collision with the AV at $t_c$, and includes a safe path from $t_0$ 
to $t_{end}=t_c+1$ such that the AV can avoid the collision, but there are no safe 
paths at $t_0+1$. Critical scenarios are interesting in that they include a collision
with the AV and also a safe trajectory to avoid the collision.

In this approach, each node can store the state of all vehicles independently of 
other nodes. This would allow actors to follow an active policy, and react to 
actions taken by the AV during scenario characterization, which enumerates all 
possible trajectories for the AV in the scenario. Each trajectory representing 
the AV states will then store the state of other actors as they react to the 
specific action taken by the AV in that trajectory.
We implemented this tree-based approach and found it to have limited scalability, 
high storage needs, and challenging to parallelize, all of which are major practical limitations.

Alternatively, we can fix the trajectories of all actors other than the AV (each actor follows a passive policy), and obtain a more efficient implementation based on tensor representations. 
This preferred tensor-based approach quantizes the space, speed, and orientation of the AV. In this representation, a 4D tensor (where dimensions are coordinates $(x,y)$, speed $v$ and orientation $\theta$) keeps track of all possible valid states for the AV at $t_i$ represented by cells, as shown in Fig.~\ref{fig:tensor}(a).
For each valid state (marked in green) at $t_i$, the next reachable states at $t_{i+1}$ are computed as explained before, by calculating the annulus sector in Fig.~\ref{fig:quant} 
and quantizing the AV state in those cells including the coordinates, orientation and speed of the AV. 
To determine next valid states from all reachable states, we need to check each state with road boundary and collision with other actors, determined by actor states at $t_{i+1}$. The next AV states that are valid are then marked in the state tensor at $t_{i+1}$ as shown in Fig.~\ref{fig:tensor}(a).
We speed up the computation of valid states from reachable states
by precomputing a map matrix for the scenario and a collision tensor for each $t_i$ in the scenario as shown in Fig.~\ref{fig:tensor}(b) to store invalid states for the AV.
The map matrix stores the invalid coordinates ($x$, $y$) based on road boundaries, and the collision tensor stores the state of AV ($x$, $y$, and $\theta$) that would be in collision with other actors based on their trajectory.
As mentioned before, the tensor method assumes that the trajectories of other actors are fixed (either taken from real data or as obtained from simulation based on the trajectory taken by the AV). Therefore, each actor's occupancy is known beforehand and can be used to compute the collision tensor.
Compared to the tree approach which stores the state of all actors in a node, the tensor approach only stores the state of AV in the 4D state tensor and assumes a fixed trajectory for other actors, stored as the collision tensor. %The tensor approach is therefore unable to capture active policy for other actors. 
We discuss the distinctions between tensor-based and tree-based approaches further in Section~\ref{sec:discussion}.

The propagation of AV state tensor is repeated for all time-steps, and 
the final tensor at $t_{end}$ gives the total number of safe paths for that scenario. Each cell in the final tensor has a non-negative integer value, indicating the number of safe trajectories which end in that state for time-step $t_{end}$ (if two trajectories converge to a state, the cell value would be $>1$). Therefore, the summation of all the cells in the final tensor is equal to the total number of safe paths.
We define the metrics in the next section, with the first metric computed directly from the number of safe paths, and the rest computed using a similar propagation strategy in the tensor space.

\subsection{Characterization Metrics}\label{sec:metrics}

We define the following set of metrics to characterize a given scenario with the 
last metric (CriticalTime) only defined for critical scenarios that include a 
collision with the AV in the sequence.
These metrics are defined such that a higher metric value indicates a more challenging scenario.

\begin{itemize}
   \item \textbf{SafePathInv ($1/\#p$):} Inverse of the number of safe paths ($\#p$) available to the AV until $t_{end}$.
   %(or one time-step beyond $t_c$). 
   When $1/\#p$ is one, the AV is constrained to follow a single trajectory to stay safe on the road. As $1/\#p$ tends to zero, the AV has many ways to avoid the accident.  
   %, and there are no alternative trajectories to avoid the collision; for a simpler scenario, $1/\#p$ tends to zero.
   \item \textbf{UnsafePercent ($p_c\%$):} The percentage of paths leading to a collision within a given scenario, among all the paths that do not lead the AV off-road, in absence of other vehicles.
   \item \textbf{AvgEffort ($E[e_s]$):} We compute the AV effort to navigate 
   %from a parent to a child node 
   from one cell to the next in state tensor
   as the sum of the absolute values of the steering and acceleration controls. To compute the effort needed to navigate a safe path ($e_s$), we add effort values of all the cells in the path. We compute the average effort for all the safe paths found to obtain AvgEffort ($E[e_s]$). A large value indicates that, on the average, high-effort navigations (i.e. complex maneuvers) are needed to avoid the accident.
   \item \textbf{MinEffort ($\min[e_s]$):} The minimum effort required to navigate through a safe path. If MinEffort is zero, the AV does not have to take any action to avoid the collision.
   \item \textbf{NarrowInv ($1/E[\min[c_s]]$):} We 
   define a branching factor for each cell at time $t_i$ as the number of cells that can be reached in the tensor at time $t_{i+1}$ from that cell, and
   measure how narrow a safe path is by computing the minimum number of branches for each cell on a safe path ($s$). We refer to it as $\min[c_s]$, where $c_s$ indicates the number of branches of a cell along the path. $E[\min[c_s]]$ is the average of the minimum number of branches for all safe paths; its inverse (referred to as NarrowInv) is one (or close to one) when the AV has only one (or few) options to navigate through a safe path. A narrow path suggests that an evasive maneuver requires precision during execution. % ; on the other hand, it tends to zero when even an inaccurate execution of the evasive maneuver leads to safe finale.
   \item \textbf{CriticalTime ($t_{critical}$)}: For critical scenarios, the minimum time (in seconds) required before $t_c$ such that the accident can be avoided by an AV. A large $t_{critical}$ value implies that the AV should take an action early to avoid the collision.
\end{itemize}

% To compute these metrics, 
Each metric is assigned a corresponding tensor and is computed using tensor propagation, i.e., propagating the metric through the tensor at each time-step from $t_0$ to $t_{end}$.
SafePathInv is directly computed from the total number of safe paths as explained before, by summing all the cells in the final state tensor
 at $t_{end}$.
Similarly, a tensor is assigned to on-road paths
in which each cell keeps track of the number of all paths reachable to that cell which are on the road (and can potentially be in collision with other actors).
In propagating this tensor at each time-step, only the map matrix is used and the collision tensor is not considered.
We compute the second metric, UnsafePercent, from the 
total number of safe paths (obtained from the first tensor) and the total number of on-road paths (obtained from the second tensor).

The tensor corresponding to AvgEffort stores at each cell (corresponding to a state) the effort required to move from the previous state to the current state.
If two or more trajectories converge in one state, the effort for that cell is computed as the sum of all efforts from prior states. To compute AvgEffort, the values of all the cells for the tensor at $t_{end}$ are added and then divided by total number of safe paths computed previously.
The tensor for MinEffort propagates the effort for each trajectory, and when two or more trajectories converge in one state, picks the smallest value to update the corresponding cell. The minimum effort of all the safe paths is then picked from the tensor at time-step $t_{end}$ as MinEffort.
The NarrowInv tensor propagates the minimum number of branches for each cell on a path.
Therefore, if a future cell along the path has fewer branches, that number will be propagated, otherwise the current state number will be propagated. We obtain the average by summing all the values in the final tensor divided by total number of safe paths.
To find CriticalTime for critical scenarios, we compute the number of safe paths from one time-step (0.5s) before collision time $t_c$. If a safe path is found, critical time is one time-step. Otherwise, we compute the number of safe paths from two time-steps before collision time and so on until a safe path is found. The corresponding time-step would be the CriticalTime, which is the latest time-step the AV can take an action to avoid collision. 

Creating metrics for scenario characterization allows us to compute a difficulty score for scenarios in testing, as well as to measure their diversity for creating non-trivial datasets. The definition of a distance function between two scenarios, e.g., based on the characterization metrics introduced in our paper, may be useful to guarantee diversity in training and testing sets, but it is left for future investigation.

\subsection{Generating Unsafe Scenarios}\label{sec:gen}

In a simulated driving sequence, environment and states of actors at any time $t$ can be used to seed a new scenario. An uninformed selection of the vehicles' driving policies as well as a random pick of the seeding time $t$ rarely leads to the identification of interesting scenarios that are challenging for the AV. We propose to assign the policy of at least one vehicle in the scenario, such that the probability of an accident with the AV increases.

We use NVIDIA DRIVE Sim~\cite{DriveSim}, which enables photo-realistic driving simulation and integration of NVIDIA's AV software stack for testing perception and planning modules. While we use DRIVE Sim, the techniques described here can be applied on any driving simulator. 
A sequence is simulated in DRIVE Sim by spawning an AV and other vehicles on the map at random initial positions and with random velocities, and assigning to each vehicle (including the AV) a simple but realistic driving policy (for freeway driving). This policy can be described by a simple Markov chain where vehicles continue to drive on the same lane, or make a maneuver (with a probability $0.5$). A maneuver involves changing lane (with probability $0.6$) or speed (with probability $0.4$). Once a maneuver is initiated, the actor attempts to complete it and continues to drive on the lane unless a collision with another vehicle is imminent and an evasive maneuver (breaking or steering adjustment) is executed.

To create an unsafe scenario with high probability of collision, we first generate an unsafe sequence.
Starting from the initial sequence, we change the policy of a vehicle that is closest to the AV after $3$ seconds from the beginning of the sequence. With probability $0.5$, we select and change the policy of the second closest vehicle. For a subsequent period that is randomly selected between 3 to 5 seconds, we allow the selected vehicles to violate safety
and move them towards the AV to increase the chance of a collision.  
We refer to such a vehicle as an \emph{attacking} vehicle and indicate it with $w_{AT}$.
With some abuse of notation, we define a subset of the state of $w_{AT}$ as $\mathbf{s_{AT}}(t)=[v_i\; \theta_i]$; the attacking policy defines the rate of change of $\mathbf{s_{AT}}$ as  $\frac{d\mathbf{s_{AT}}}{dt} =[a_{AT}\; \frac{v_{AT}\tan(\delta_{AT})}{L_{AT}}]$, where $a_{AT}$ and $\delta_{AT}$ are the the acceleration and steering angle of $w_{AT}$, whereas $L_{AT}$ is the wheelbase.
We want the attacking policies to minimize the distance between the AV and attacking vehicle at positions $\mathbf{x_{AV}}$ and  $\mathbf{x_{AT}}$, respectively. 
% (Fig.~\ref{fig:attack_state}).
So, the objective is to minimize the cost function $\gamma =  \parallel\bm{x_{AT}} - \bm{x_{AV}} \parallel^2_2$. We can achieve this by %To achieve this, we find the control policy for the unsafe actor which generates a trajectory that minimizes our objective function ($\gamma$) defined as:
%$$
%\gamma_{AE} = D^2_{AE} = \parallel \bm{x}_A - \bm{x}_E \parallel^2_2% = (x_{iA}-x_{iE})^2 + (x_{jA}-x_{jE})^2
%$$
decreasing $\gamma$ at each time-step or by setting the constraint $\frac{\partial \gamma}{\partial t}< 0 $. We have:
%
%\begin{wrapfigure}{r}{0.4\textwidth}
\begin{figure}[t]
\centering
\includegraphics[width=.48\textwidth]{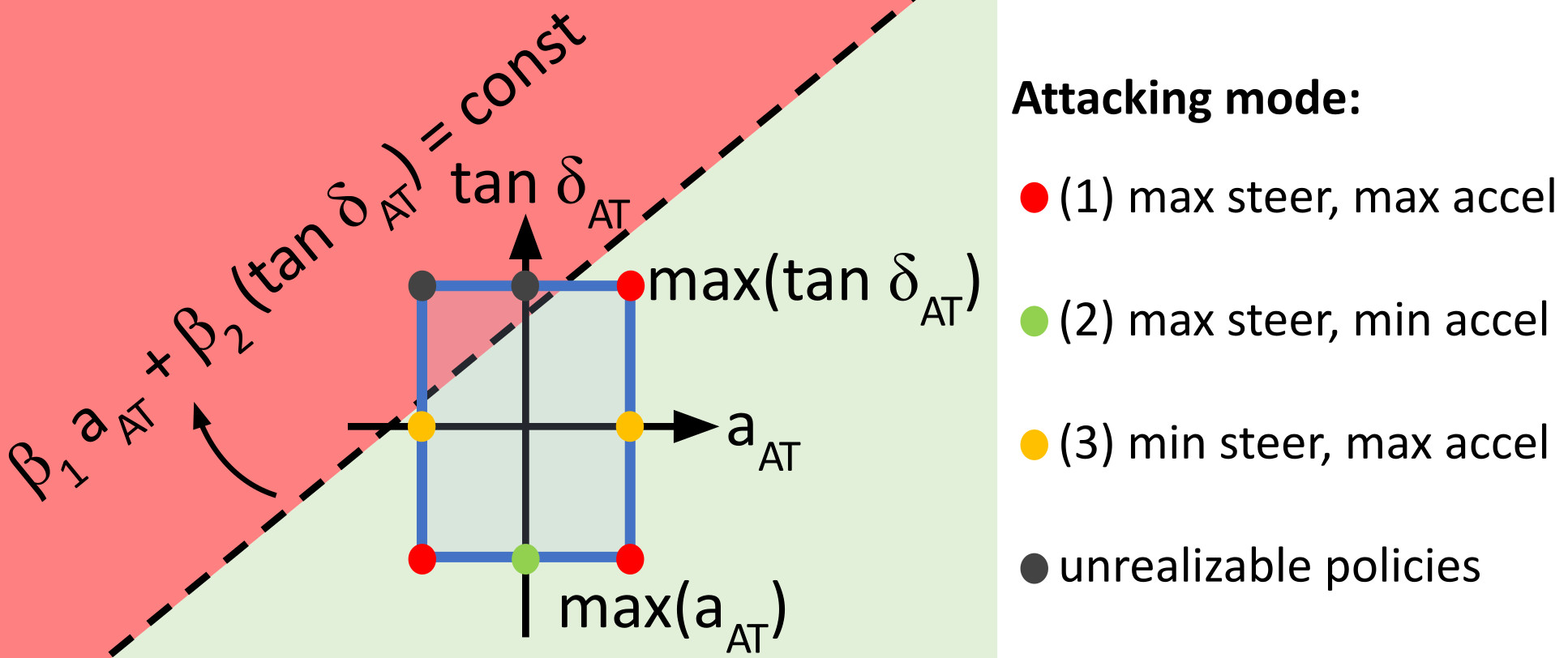}
\caption{The constraint to minimize the distance between the AV and attacking vehicle divides the $(\tan(\delta_{AT}), a_{AT})$ plane into attacking (green) and non-attacking (red) regions. Within the attacking region, we use different attacking modes. 
%\iuri{isn't red the attacking area, and green the non attacking one, with the line being neutral? I would not use valid and not valid, but attacking and non attacking if so.}
}
\label{fig:constraint}
\end{figure}
\begin{equation}
   \frac{\partial \gamma}{\partial t} = \frac{\partial \gamma}{\partial \mathbf{s_{AT}}}\frac{\mathrm d \mathbf{s_{AT}}}{\mathrm d t} + \frac{\partial \gamma}{\partial \mathbf{s_{AV}}}\frac{\mathrm d \mathbf{s_{AV}}}{\mathrm d t} < 0,
\label{eq:deriv}
\end{equation}
%
\begin{comment}
We can write
\begin{multline}
\frac{\partial \gamma_{AE}}{\partial s_A} = 2(x_{iA}-x_{iE})[\sin(\theta_A)\Delta t \quad v_A\cos(\theta_A)\Delta t]\\
+2(x_{jA}-x_{jE})][\cos(\theta_A)\Delta t \quad -v_A\sin(\theta_A)\Delta t]
\end{multline}
and derive a similar expression for $\frac{\partial \gamma_{AE}}{\partial s_E}$. Plugging these expressions into Equation~\ref{eq:deriv} generates a control constraint on the unsafe actor as following:

\begin{multline}
(\Delta x_i [\sin(\theta_A)\quad v_A\cos(\theta_A)] + \Delta x_j [\cos(\theta_A) \quad -v_A\sin(\theta_A)]).\Big [\begin{array}{c}
			a_A\\
			v_A\tan(\delta_A)/L_A
		\end{array}\Big ] + \\
(\Delta x_i [-\sin(\theta_E)\quad -v_E\cos(\theta_E)] + \Delta x_j [-\cos(\theta_E) \quad v_E\sin(\theta_E)]).\Big [\begin{array}{c}
			a_E\\
			v_E\tan(\delta_E)/L_E
		\end{array}\Big ] < 0
\end{multline}
\end{comment}
%
which can be simplified as: 
\begin{equation}
\beta_1 a_{AT} + \beta_2 \tan(\delta_{AT}) < k,
\label{eq:ctrl_constraint}
\end{equation}
where the constants $\beta_1$, $\beta_2$, and $k$ are function of the current position, speed and orientation of $w_{AV}$ and $w_{AT}$.  

\begin{comment}
\begin{wrapfigure}{r}{0.45\textwidth}
%\begin{figure}[h]
\centering
\includegraphics[width=0.35\textwidth]{figs/control.png}
\caption{The 2D control space for the unsafe actor and constraints on steering and acceleration. Any point in the triangle region which satisfies all constraints can be picked as the control policy of the unsafe actor. We present $4$ different methods (indicated by $m1$ through $m4$) with varying effort in steering and acceleration.}
\label{fig:control}
\end{wrapfigure}
\end{comment}

The attacking policy for actor $w_{AT}$ is thus defined by steering angle $\delta_{AT}$ and acceleration $a_{AT}$. In the 2D space $(\tan(\delta_{AT}), a_{AT})$, Eq.~\ref{eq:ctrl_constraint} defines an attacking hemispace (Fig.~\ref{fig:constraint}).
Additionally, we put constraints on the maximum values of steering angle and acceleration (blue rectangle in Fig.~\ref{fig:constraint}).
Since any point in space $(\tan(\delta_{AT}), a_{AT})$ that satisfies the above constraint can be used as a control command for $w_{AT}$, we propose three different \emph{attacking modes}:
(1) maximum acceleration and steering, or maximum effort, (2) minimum acceleration and maximum steering, and (3) minimum steering and maximum acceleration.
These policies are obtained by selecting the corresponding point on the blue rectangle in Fig.~\ref{fig:constraint}.
%The fourth policy (minimum effort) is obtained by selecting the point in the attacking hemispace closest to the origin. \zahra{we don't implement this}
Note that depending on the constraint line defining the attacking hemispace, not all attacking policies might be available or realizable in practice for all scenarios. These policies that fall in the non-attacking hemispher are also shown in Fig.~\ref{fig:constraint} as \emph{unrealizable policies}.

These controllers can be used to control the aggressiveness of the attack, and ensure that the generated adversarial scenarios are more realistic. For example, maximum braking can simulate a driver who suddenly brakes in reaction to something they see, or maximum steering can be set to simulate a distracted driver drifting to the next lane.
Additionally, we can seed this generator from real driving scenarios.
Starting from a recorded real-world scenario, we can modify the behavior of one driver for 3 seconds to generate an incident. It is worth noting that even if this may be a pessimistic scenario, it is still acceptable for benchmarking as it can compare the performance of different AV systems without requiring that all scenarios must be safely navigated.

To create a critical scenario for characterization, we freeze the trajectories of all the vehicles except for the AV and attacker(s) in the sequence up to $t_c$.
Since the attacker can abruptly stop because of the accident at $t_c$ in the sequence, we go back in time one step and extrapolate the trajectory to continue with the same speed after the accident.
The scenario is created by moving back in time from $t_c$ until we find the time $t_0$ such that the AV can find at least one safe path to avoid the collision.

%\subsection{Critical Scenario Generation for Characterization} 
\begin{comment}
\begin{figure}
    \centering
    \subfigure[]{\includegraphics[width=0.35\textwidth,clip,trim=3cm 0cm 1cm 1cm]{./figs/fig2-a.PNG}
    \label{fig:egotree}}
    %\hspace{1cm}
    \subfigure[]{\includegraphics[width=0.35\textwidth]{./figs/fig2-b.png}\label{fig:egograph}}
    %\hspace{1cm}
    %\subfigure[]{\includegraphics[width=0.3\textwidth]{./figs/ego_steps.png}\label{fig:egosteps}}
    \caption{Scenario scoring based on search for safe driving policy. (a) shows the quantized region on the map where the AV can travel in the next time-step based on maximum allowed steering and acceleration. (b) is the tree representation of cell locations for the AV in different time-steps during the scenario roll-out. Cells marked red represent the invalid locations which intersect with other actors in the scenario and indicate a collision.}
    \label{fig:scenario-search}
\end{figure}
\end{comment}

\section{Evaluation}\label{eval}

\begin{comment}
\begin{table*}[]
\centering
\resizebox{0.8\textwidth}{!} {
\begin{tabular}{|l
|c|c|c|}
\hline
%\multicolumn{1}{|c|}{\multirow{2}{*}{Attacker constraints:}} & \multicolumn{3}{c|}{Attacking mode} \\ \cline{2-4} 
\multicolumn{1}{|c|}{Attacker constraints:} & \multicolumn{3}{c|}{Attacking mode} \\ \cline{2-4} 
\multicolumn{1}{|c|}{Steering, acceleration ranges} & \multicolumn{1}{l|}{Max. steer, max. accel} & \multicolumn{1}{l|}{Max. steer, min. accel} & \multicolumn{1}{l|}{Min. steer, max. accel} \\ \hline
Max. steer=0.2, accel=0.8 & 37 & 36 & 34 \\ \hline
Max. steer=0.1, accel=0.4 & 26 & 26 & 23 \\ \hline
Max. steer=0.2, accel=0.1 & 9 & 9 & 8 \\ \hline
\end{tabular}
}
\caption{Number of accidents in scenarios generated for different attacker constraints (rows) and attacking modes (columns) are shows here. 
%
}
\label{tab:modes}
\end{table*}
\end{comment}

\begin{figure}
\centering
\includegraphics[width=.43\textwidth]{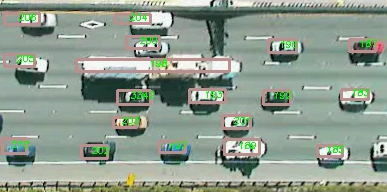}
\caption{Sample image of video data from NGSIM showing vehicle positions along with vehicle identification numbers and extracted vehicle dimensions.}
\label{fig:ngsim}
\end{figure}

\begin{figure*}
\centering
\subfloat[]{\includegraphics[width=0.5\textwidth]{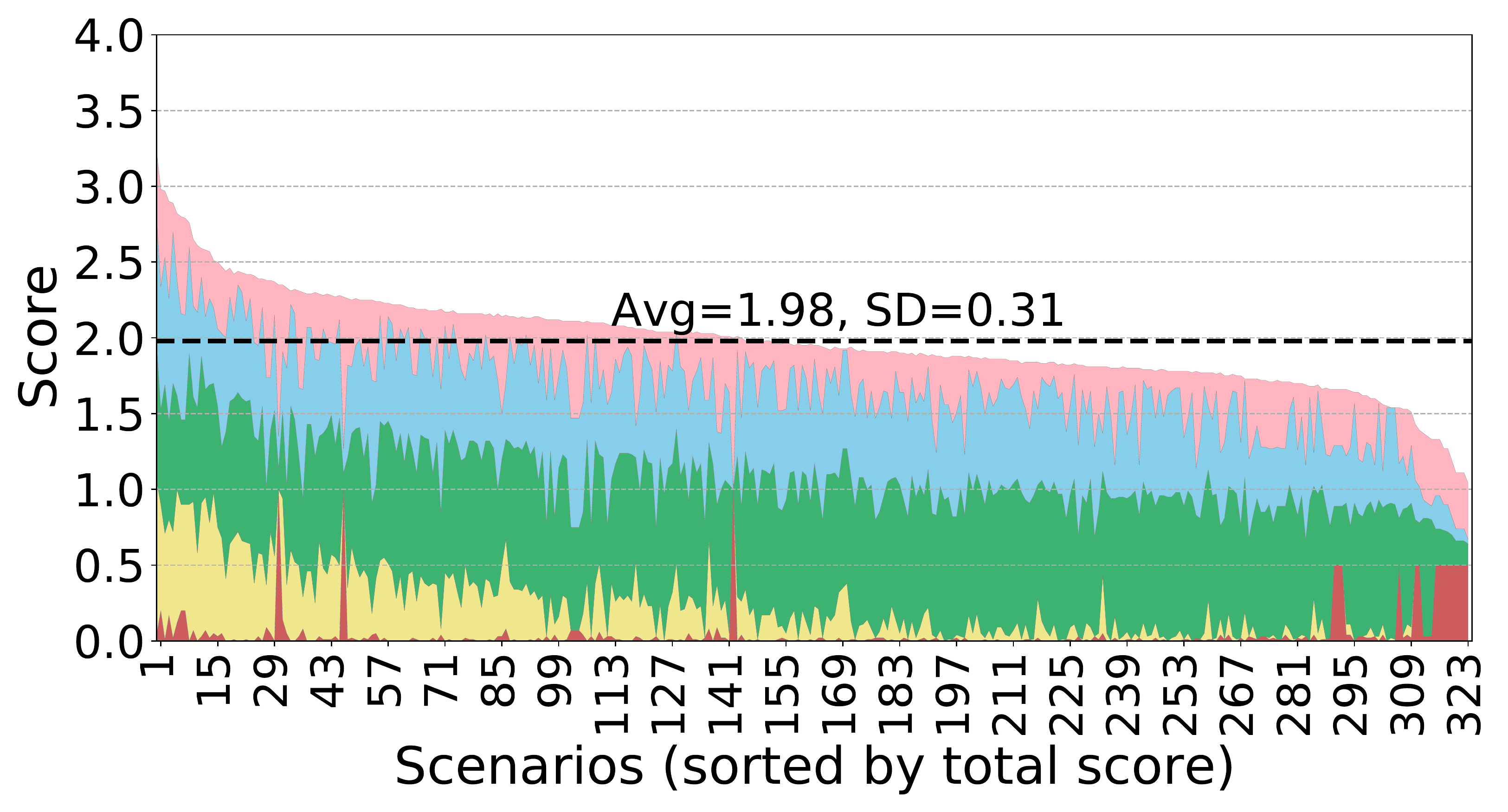}\label{fig:ngsim_score}}
\subfloat[]{\includegraphics[width=0.5\textwidth]{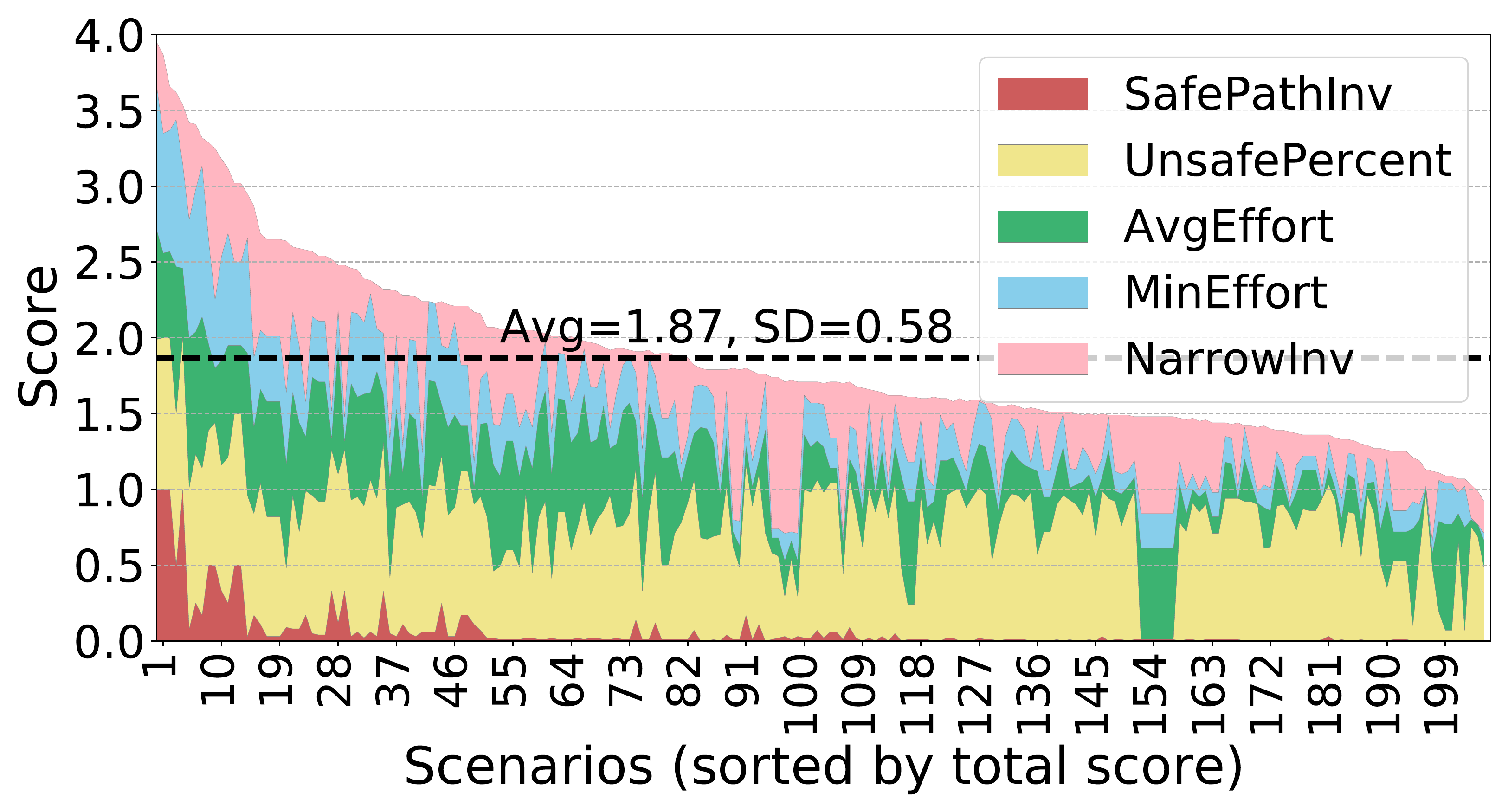}\label{fig:adv_score}}
\caption{Score and metric break down of (a) scenarios from NGSIM, and (b) generated adversarial scenarios with accidents. While the average scores are close between NGSIM and adversarial scenarios, the standard deviation for the set of adversarial scenarios is higher, indicating a more diverse set in terms of total score.}
\label{fig:metrics}
\end{figure*}

We use our metrics to characterize real-world driving scenarios extracted from the Next Generation Simulation (NGSIM)~\cite{ngsim} dataset, as well as scenarios generated by our method described in Section~\ref{sec:gen} that creates adversarial (unsafe) scenarios with attacking actors.

\subsection{NGSIM Scenario Characterization}
We use the dataset in NGSIM which provides vehicle trajectory data on the Interstate 80 (I-80) Freeway in the San Francisco Bay area covering a 500 meter length area, and a total of 45 minutes.
Fig.~\ref{fig:ngsim} shows a snapshot of the video data in NGSIM. The vehicle trajectory data includes the coordinates, size, velocity, acceleration, and type (motorcycle, automobile, and truck) for each vehicle. 
We extracted 3 second scenarios from the trajectory data between 4:00pm to 4:15pm and picked a random vehicle as AV each time.

We compute 5 characterization metrics for the extracted scenarios (omitting CriticalTime for non-critical scenarios of constant length). 
We assign a \emph{score} as the sum of the metrics, after normalization based on the range of values over a set of scenarios. 
For normalizing each metric, each value is subtracted with the minimum in set and then the difference is divided by the range of that set.
This normalization is done across the set of scenarios being compared.
The break down of our metrics for scenarios extracted from NGSIM 
is depicted in Fig.~\ref{fig:metrics}, sorting scenarios by the score. Total scenario score can also be extracted from this graph by reading the aggregate score, and the average and standard deviation are computed over the entire set of scenarios.

NGSIM contains crowded highway data at rush hour, with vehicles that can only make little movement at some instances of time, and a lot of braking is required to avoid collision.
As can be observed from this plot, this translates into large AvgEffort and MinEffort for navigating the extracted scenarios.
Overall, the NGSIM scenarios are homogeneous with a small standard deviation for total score and variation in metrics due to similar driving conditions for recorded data.

\begin{table}[]
\centering
\begin{tabular}{lccc}\toprule
\multirow{2}{*}{\begin{tabular}[c]{@{}l@{}}Attacker constraints:\\ Steering (S), Acceleration (A) \end{tabular}} & \multicolumn{3}{c}{Attacking Mode}\\ 
\cmidrule(lr{0.5em}){2-4}
& \begin{tabular}[c]{@{}c@{}}Max S\\ Max A\end{tabular} & \begin{tabular}[c]{@{}c@{}}Max S\\ Min A\end{tabular} & \begin{tabular}[c]{@{}c@{}}Min S\\ Max A\end{tabular} \\
\midrule
Max steering=0.2, acceleration=0.8 & 37 & 36 & 34 \\
Max steering=0.1, acceleration=0.4 & 26 & 26 & 23 \\
Max steering=0.2, acceleration=0.1 & 9 & 9 & 8 \\      
\bottomrule
\end{tabular}
\caption{Number of accidents in scenarios generated for different attacker constraints and attacking modes. 
}
\label{tab:modes}
\end{table}

\subsection{Adversarial Scenario Characterization}
To characterize adversarial scenarios using
our characterization method, we create $70$ random initial sequences
in DRIVE Sim. For each of them, we generate $9$ sequences with our unsafe scenario generation algorithm using three attacker constraints (limiting attacker acceleration and steering to a maximum amount)
and three attacking modes, as described in Table~\ref{tab:modes}. %, as described in the previous section. 
We use the intelligent driver model for the AV~\cite{treiber2000congested}. Table~\ref{tab:modes} shows the number of accidents we observe out of a total of $630$ sequences. Generating these scenarios takes  less than 3 hours on a system with an Intel Core i7-7800X CPU, 2 NVIDIA RTX 2080 GPUs, and 32GB memory. 
Using these configurations, we could create $208$ accidents with the AV vehicle.
Table~\ref{tab:modes} shows the breakdown for each attacking mode. As expected, increasing the maximum acceleration and steering limits leads to a higher number of accidents
(first row). We also observe that attacking with the minimum steering mode is less effective (last column).

\begin{comment}
\begin{figure}
    \centering
    \includegraphics[width=\textwidth]{./figs/scores_narrow.png}
    \caption{Score of the 208 unsafe scenarios with accident and its break down in six metrics}
    \label{fig:metrics}
\end{figure}
\end{comment}

\begin{comment}
\begin{figure*}
    \centering
    \includegraphics[width=\textwidth]{./figs/scenario-results.png}
    \vspace{-0.25in}
    \caption{Score and metric break down of (a) scenarios from NGSIM, and (b) generated adversarial scenarios with accidents.}
    \vspace{-0.2in}
    \label{fig:metrics}
\end{figure*}
\end{comment}

\begin{figure*}
\centering
\subfloat[]{\includegraphics[width=0.17\textwidth]{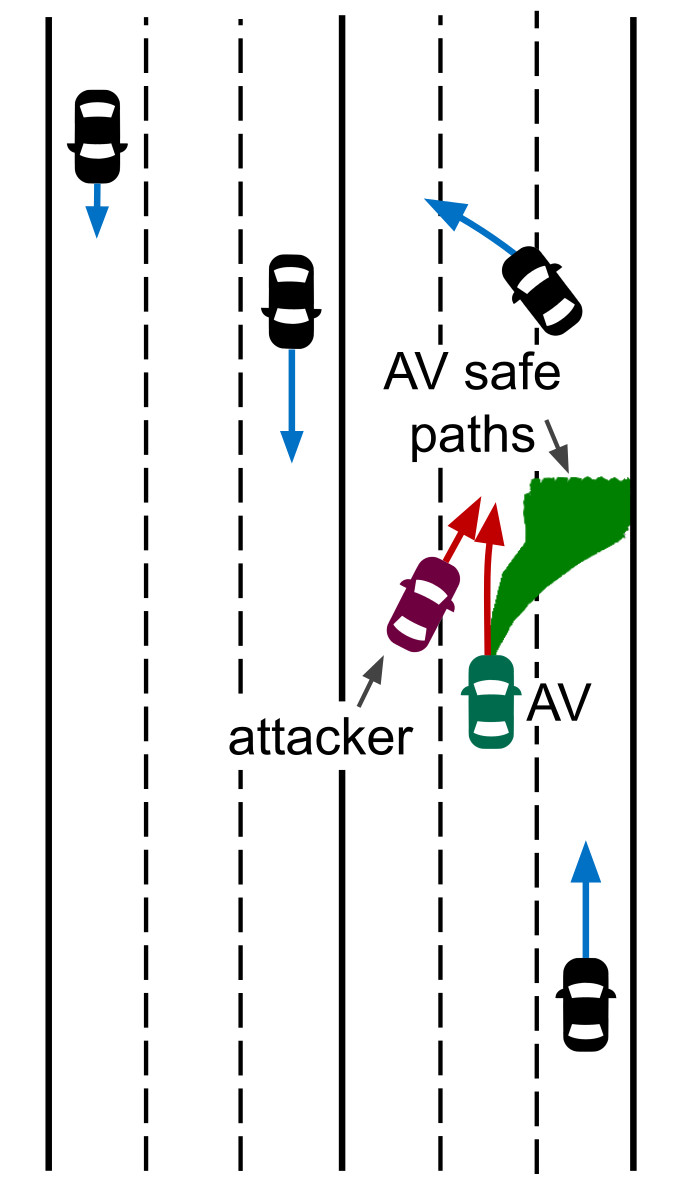}}\hfill
\subfloat[]{\includegraphics[width=0.17\textwidth]{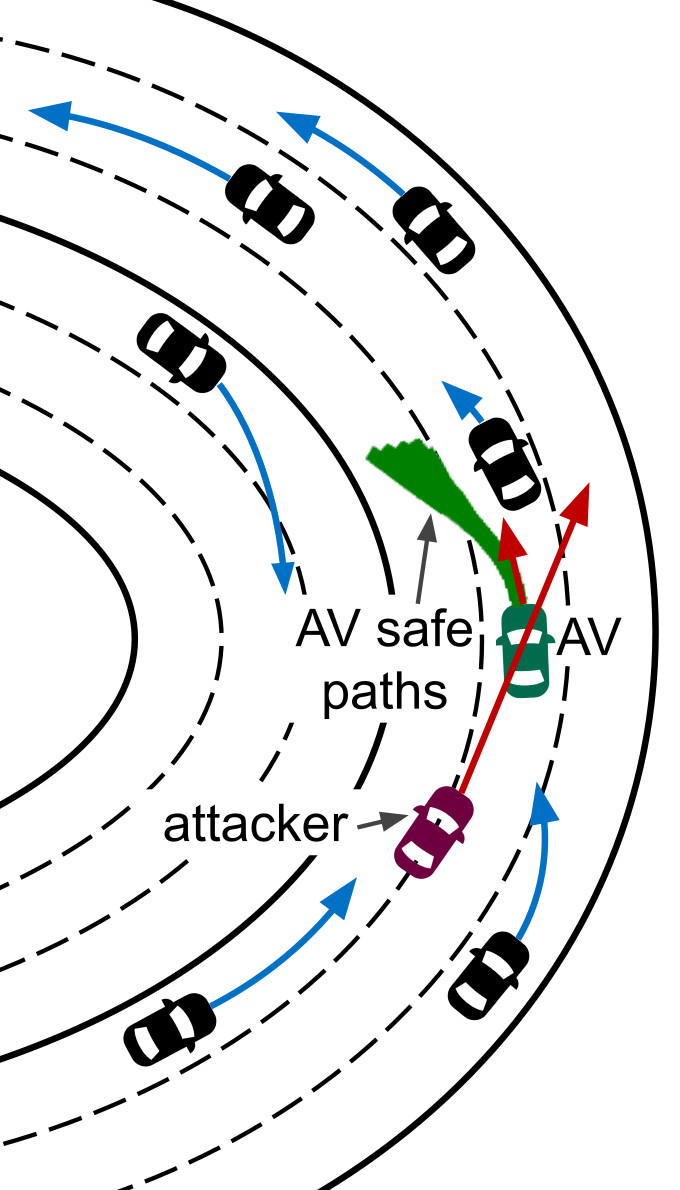}}\hfill
\subfloat[]{\includegraphics[width=0.17\textwidth]{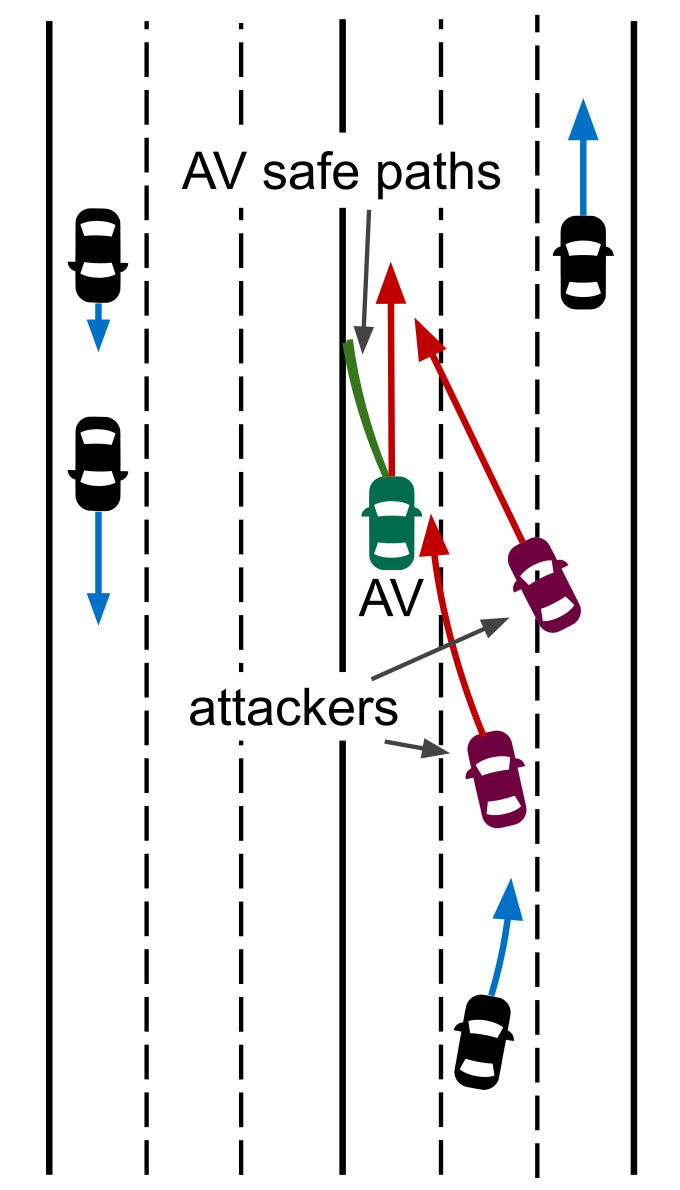}}\hfill
\subfloat[]{\includegraphics[width=0.4\textwidth]{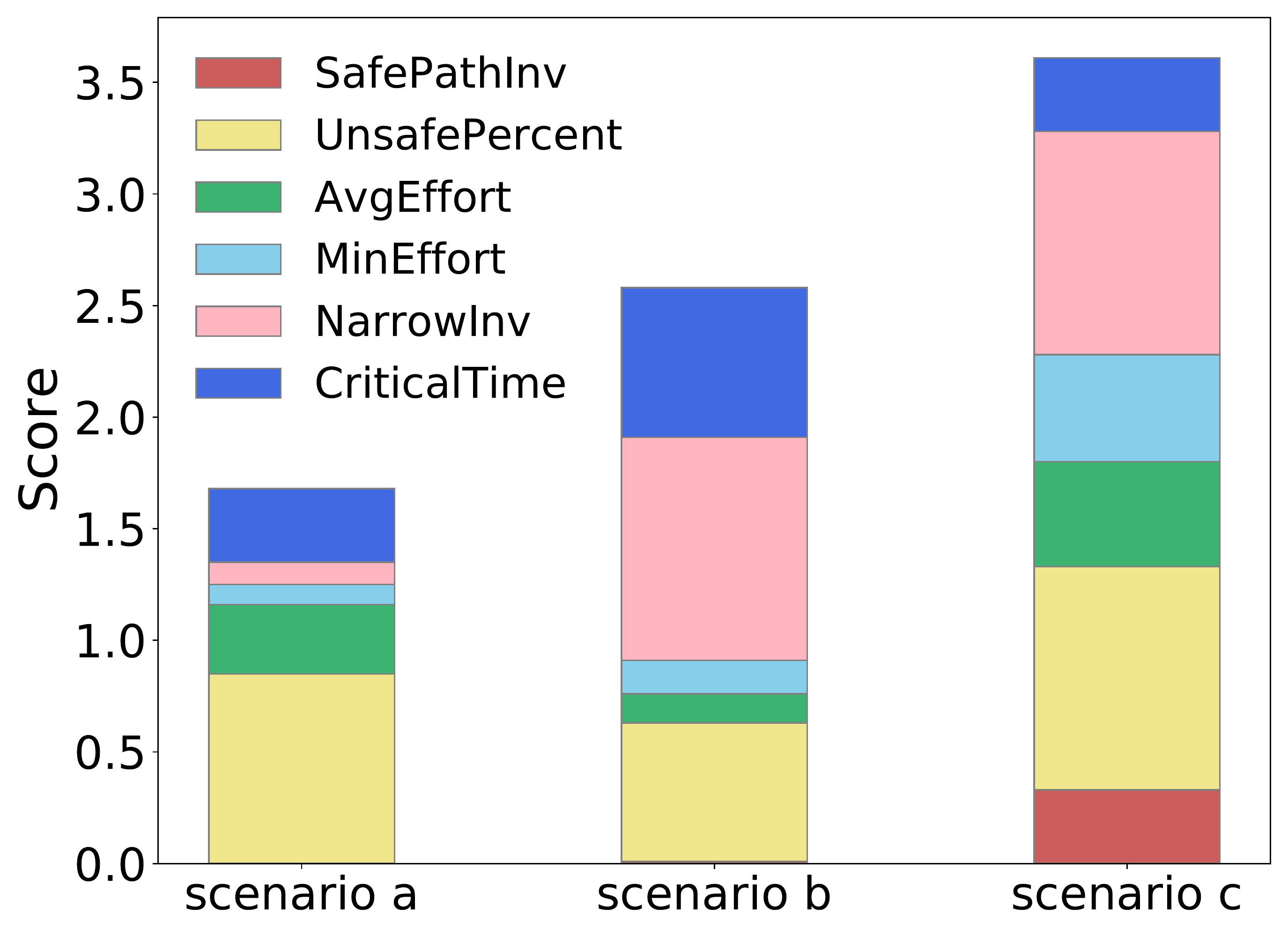}\label{fig:ex_stacked}}
\caption{Examples of generated scenarios with low (a, %$1.32$
$1.68$), medium (b, $2.58$), and high (c, $3.61$) scores. Panel (d) shows the values of the six metrics for each of them.}
\label{fig:three_scenarios}
\end{figure*}

Similar to scenarios extracted from NGSIM, we compute 5 metrics for adversarially generated scenarios as shown in Fig.~\ref{fig:metrics}(b).
For these scenarios,
SafePathInv is large only for the most challenging scenarios ($1-17$ in the graph) which have an extremely small number of safe paths.
For the same scenarios, MinEffort is consistently high, indicating the effort required to navigate through the available safe paths is high.
Additionally, there are few possible trajectories to avoid the collision and all of them require a significant number of evasive maneuvers.
NarrowInv is also high for the most challenging scenarios ($1-73$), suggesting that
the safe paths require at least one precise maneuver for navigation.
AvgEffort is large for both the challenging ($1-17$) and easy to solve ($177-218$) scenarios. For the challenging scenarios, few safe paths with a complex geometry are present, which makes the average effort high. For easy scenarios, on the other hand, the existence of many safe paths (including the complex ones) makes the average effort high.
UnsafePercent is almost constant for scenarios $1-150$ and then decreases, indicating that any on-road trajectory followed by the AV is a safe for the low-scored scenarios.

As can be observed from Fig.~\ref{fig:metrics}, the NGSIM scenarios have slightly higher average score, but adversarial scenarios have higher standard deviation in total score. This indicates more variety in scenario difficulty for adversarial scenarios with different attacking modes, compared to more homogeneous real-world highway driving.
The most difficult scenario in NGSIM has a score of 3.22, whereas the highest score in adversarial scenarios is 3.95. 
Adversarial scenarios have higher UnsafePercent for all scenarios, which is explained by the presence of the attacking vehicle(s). On the other hand, NGSIM scenarios are very similar on the AvgEffort and MinEffort metrics.

\begin{comment}
\begin{figure*}
    \centering
    \includegraphics[width=\textwidth]{figs/examples_narrow.png}
    \caption{Examples of generated scenarios with low (a, $1.32$), medium (b, $2.58$), and high (c, $3.61$) scores. Panel (d) shows the values of the six metrics for each of them. \iuri{indicate the attacking car in some way?}}
    \label{fig:three_scenarios}
\end{figure*}

\begin{figure*}
\centering
\subfloat{\includegraphics[width=0.6\textwidth]{./figs/examples_scenarios.png}\label{fig:ex_scenarios}}
\subfloat{\includegraphics[width=0.4\textwidth]{./figs/scores_stacked.pdf}\label{fig:ex_stacked}}
\caption{Examples of generated scenarios with low (a, $1.32$), medium (b, $2.58$), and high (c, $3.61$) scores. Panel (d) shows the values of the six metrics for each of them. \iuri{indicate the attacking car in some way?}}
\label{fig:three_scenarios}
\end{figure*}
\end{comment}

\subsection{Characterization Metrics Intuition}
To better demonstrate the effectiveness of the proposed set of metrics
in guiding the selection of challenging scenarios for safety testing, we sample and analyze three generated scenarios with low, medium, and high scores.
For these scenarios, we compute the total score using our 6 characterization metrics.
Intuitively, scenario (a) in Fig.~\ref{fig:three_scenarios} is a simple scenario where the AV can avoid the car attacking from the left by moving 
to the open space in the right lane (small SafePathInv).
While the AvgEffort is not minimal due to the variety of the possible safe paths, the MinEffort is small, suggesting that the AV can follow a simple trajectory to solve the scenario.
Scenario (b) is comparatively more challenging with an attacker blocking the lane in front and the other attaching the AV from the back.
In this case, the safe driving paths exist by driving the AV to the left. 
CriticalTime for this scenario is higher compared to Scenario (a), suggesting an action has to be taken earlier to avoid collision. A high NarrowInv also suggests that most of the available safe paths require a precise maneuver for navigation.
Finally, Scenario (c) is the most challenging scenario among the three, where two attackers are moving behind and in front of the AV. Most of the paths are unsafe (high UnsafePercent), and the road boundary limits the total number of paths (high SafePathInv). 
 Fig.~\ref{fig:three_scenarios}d shows the metrics of the three scenarios, which allows an immediate multi-dimensional comparison of their characteristics. The height of the stacked bar shows the score.

\begin{comment}
Similarly, extracted scenarios with low and high scores are depicted in Fig.~\ref{fig:ngsim_scenarios} a and b respectively. In the scenario with a low score, the selected ego vehicle has a clear path in the front, whereas the high score scenario depicts a congested road with several vehicles blocking the ego vehicle which could lead to a potentially dangerous condition.

\begin{figure}
\centering
\subfloat[]{\includegraphics[width=0.21\textwidth]{figs/ngsim_low.png}}
\hfill
\subfloat[]{\includegraphics[width=0.21\textwidth]{figs/ngsim_high.png}}
\caption{Examples of scenarios extracted from NGSIM with low and high scores}
\label{fig:ngsim_scenarios}
\end{figure}
\end{comment}

\subsection{Limitations, Applications, and Future Directions}\label{sec:discussion}

To generate an unsafe scenario with our scenario generation method, we change the policy of the attacker(s), while making no alterations to the driving policies of other vehicles. It is possible that other vehicles would react to the changed behavior of the attackers.
Prior research proposed methods to generate traffic based on real data, including reactive actors models~\cite{henaff2019model} which can be combined with our unsafe scenario generation as an interesting future direction.

Similarly, other vehicles including the attacker(s) can react to the change in the AV's driving policy explored during scenario characterization. In this work, when we compute metric tensors, we fix the trajectories of all the non-AV vehicles based on the initial sequence. While our tensor method makes this assumption to speed up computation and improve scalability, the tree method for computing metrics can incorporate such reactions from actors, albeit at the cost of storage and scalability.
The tree representation can store all actors' states in \emph{each} node, enabling the method to capture actor reactions to different AV policies enumerated during characterization. However, the tree approach is slow with memory consumption that grows with the number of steps, while it stays constant in the tensor approach.

The set of metrics defined in this work can be expanded.
For example, a \emph{safety} metric can be added based on distance or time to collision at each time-step. With this metric, we can search for paths that maximize safety and minimize effort. By replacing scenario trajectories with actor predictions, this metric can be used for developing a driving policy.

In addition to propagating the state tensor for computing characterization metrics, we can ``go back in time'' to place the AV in critical states that would likely result in a crash. This could be used for generating scenarios which have specific characteristics for safety testing.
We also note that propagating the metrics at each time-step can be seen as a convolution operation on the tensor, which is an operation that has been optimized
and could be leveraged for scenario characterization with our formulation.

\section{Related Work}\label{related}

There has been some progress in generating unsafe driving scenarios, fueled by recent advancements in the driving simulators and real-traffic data-sets. 
In~\cite{o2018scalable}, a base distribution representing standard traffic behavior is learned from data. % using an ensemble of generative adversarial imitation learning (GAIL) models. 
An adaptive importance sampling method is applied to learn alternative distributions from the base distribution, that can generate accidents more frequently. This method is limited to the road segments and types of scenarios present in the dataset, unlike our generation method. The scenarios are ranked based on their likelihood under the base distribution, without considering the avoidability of the accident, which is another key criteria for testing.
    % However, learning policies that generalize well to different road scenarios and effectively produce unsafe situations remains challenging.
%\iuri{possible limitation: poor control on the attack policy - is it strong or weak?}\zahra{not sure, based on traffic data}
%The toolchain presented can test an AV as a whole system, simulating the driving policy of the ego-vehicle by viewing it as a black-box model, using a photo-realistic and physics-based simulator. 
%
In~\cite{abeysirigoonawardena2019generating}, authors use Bayesian Optimization to generate adversarial scenarios that increase the risk of collision with pedestrians and vehicles. This method scales poorly with the increasing number of actors. Our greedy and gradient-based approach to generating unsafe scenario performs much faster.
    % The training set is then augmented with expert reactions to adversarial scenarios and the policy is retrained using vision-based imitation learning to obtain safer driving behavior.
    % The fact that intervention of human experts and that scenario roll-out is performed in the simulator, make this approach much slower when compared to ours.
% However, this method doesn't scale well to high dimensional settings.
%\iuri{possible limitations: it does not generalize well, retraining needed if conditions change?} \zahra{no training involved afaik}
%
In \cite{corso2019adaptive}, the authors model the problem of finding failure cases as a Markov decision process and use reinforcement learning to solve it. Methods based on reinforcement learning need long training times for each new configuration, which limits their applicability. We require no training to generate an unsafe scenario.
None of these techniques characterize generated scenario based on accident avoidability. Our method can be applied to scenarios generated by any of the above techniques.

Fault injection methods~\cite{jha2018avfi} study resilience to hardware errors, and do not focus on scenario generation or classification for AV testing. Falsification methods~\cite{dreossi2019compositional,dreossi2019verifai} introduce perturbations in a frame or configuration of a scenario to falsify a defined specification, but do not provide a method to categorize and differentiate the generated scenarios. Importance sampling~\cite{zhao2016accelerated} accelerates rare event generation based on a learnt distribution, but does not provide more detail on the event or explanation on avoiding it, critical for safe AV development. Scenic~\cite{fremont2019scenic} provides a probabilistic programming language for describing scenarios and generating them by constrained sampling, or manual definition of hard scenarios.

%Our paper also provides a way to distinguish between scenarios, not provided by any prior paper. 
To the best of our knowledge, none of prior work can compare scenarios based on whether the unsafe condition can be avoided by a narrow path and/or quick reaction to the hazard. % This allows developers to identify classes of unsafe situations, each characterized by our metrics, allowing for creation of a diverse set of test scenarios. Other works on scenario generation can benefit from our characterization method to prune or diversify their test set. 
% Siva: I think the above two sentences should belong to Section III.D if at all we should include them.
We found~\cite{chou2018using} to be the closest work to ours. In this work, Chou et. al. define interesting cases as the ones where ensuring safety is hard but not impossible, and it can also benefit from providing feedback to developers on failing case characteristics using our characterization method.

\section{Conclusion}\label{conclusion}
This paper presented a novel characterization method that quantifies the difficulty of a scenario based on several defined metrics. We characterize a scenario by enumerating possible safe paths to navigate through the scenario, and computing metrics such as the narrowness of safe paths, and the effort required to follow them. These metrics provide insights into how the AV could have avoided potential collisions, which is key to developing a safer AV. We characterize scenarios extracted from pre-recorded real world data (NGSIM), as well as adversarial scenarios that we generated. We develop a fast method to generate potentially unsafe scenarios and demonstrate that it can generate 240 potentially unsafe scenarios per hour with more than a third of the scenarios resulting in collisions on a state-of-the-art driving simulator.

\addtolength{\textheight}{-12cm}   % This command serves to balance the column lengths

%%%%%%%%%%%%%%%%%%%%%%%%%%%%%%%%%%%%%%%%%%%%%%%%%%%%%%%%%%%%%%%%%%%%%%%%%%%%%%%%

%\section*{APPENDIX}


%\section*{ACKNOWLEDGMENT}

%Put acknowledgement.

\bibliographystyle{unsrt}
\bibliography{ref}

\end{document}